# Enhancing Energy Efficiency and Reliability in Autonomous Systems Estimation using Neuromorphic Approach

Reza Ahmadvand, Sarah Safura Sharif, Yaser Mike Banad

*Abstract*— Energy efficiency and reliability have long been crucial factors for ensuring cost-effective and safe missions in autonomous systems computers. With the rapid evolution of industries such as space robotics and advanced air mobility, the demand for these low size, weight, and power (SWaP) computers has grown significantly. This study focuses on introducing an estimation framework based on spike coding theories and spiking neural networks (SNN), leveraging the efficiency and scalability of neuromorphic computers. Therefore, we propose an SNN-based Kalman filter (KF), a fundamental and widely adopted optimal strategy for well-defined linear systems. Furthermore, based on the modified sliding innovation filter (MSIF) we present a robust strategy called SNN-MSIF. Notably, the weight matrices of the networks are designed according to the system model, eliminating the need for learning. To evaluate the effectiveness of the proposed strategies, we compare them to their algorithmic counterparts, namely the KF and the MSIF, using Monte Carlo simulations. Additionally, we assess the robustness of SNN-MSIF by comparing it to SNN-KF in the presence of modeling uncertainties and neuron loss. Our results demonstrate the applicability of the proposed methods and highlight the superior performance of SNN-MSIF in terms of accuracy and robustness. Furthermore, the spiking pattern observed from the networks serves as evidence of the energy efficiency achieved by the proposed methods, as they exhibited an impressive reduction of approximately 97 percent in emitted spikes compared to possible spikes.

## I. INTRODUCTION

The design and industrial fabrication of autonomous systems onboard computers places significant emphasis on energy efficiency and reliability, particularly when considering edge computing applications in resource-constrained and dynamic environments such as space robotics [1], internet-of-things (IoT) [2], advanced air mobility (AAM) [3], and SWaP engineering [4]. In this context, neuromorphic computers present an appealing solution to address these dominant performance factors for edge intelligence [5]. Neuromorphic computers, characterized by their inherent scalability and extremely low power operation, offer significant advantages over traditional von Neumann computer architectures, operating at orders of magnitude lower power consumption [6]. Neuromorphic computers represent a new generation of computing tools that leverage neural circuits composed of neurons and synapses to perform computations. In contrast to von Neumann architectures which utilize binary data and sequential algorithmic approaches, neuromorphic computers encode data and perform computations in an event-driven and massively parallel manner, utilizing spikes, their timing, and occurrence [6, 7, 8, 9, 10]. To program a neuromorphic computer chip, the implementation of spiking neural networks (SNNs) is essential, which can be considered as a programming language for these edge devices [6]. SNNs have emerged as biologically plausible brain-inspired computing tools, bridging the gap between neuroscience and machine learning [10]. Representing the third generation of neural networks, SNNs utilize distinct network structures and neural computations that distinguish them from artificial neural networks (ANNs), leading to reduced resource requirements in terms of energy consumption, data utilization, and computational costs [10]. In addition to the energy efficiency of neuromorphic computers, their scalability can be considered as an advantage which makes them more reliable in practical applications. In contrast to traditional von Neumann architectures, where a malfunctioning component halts the entire system, neuromorphic computers exhibit a more resilient behavior. If a portion of the network becomes damaged, neuromorphic computers undergo a condition akin to neuron silencing, losing some neurons. Nevertheless, this loss can be compensated by the increased spiking rate of the remaining neurons within the network [7]. Thus, it can be concluded that in addition to energy efficiency, neuromorphic computers can be more reliable even in critical operational conditions, making them suitable for autonomous systems. From this perspective, the focus of this research is on the development of a novel neuromorphic estimation framework designed to achieve robust filtering of linear dynamical systems. This framework is specifically tailored for

Reza Ahmadvand, Sarah Sharif and Yaser Mike Banad (corresponding author) are with the Department of Electrical, and Computer Engineering, University of Oklahoma, Oklahoma, United States E-mails: *iamrezaahmadvand1@ou.edu*, *s.sh@ou.edu*, *bana@ou.edu*

implementation on neuromorphic chips, enabling its application in real-world scenarios involving autonomous robotic systems. To this aim, the introduced frameworks in [11, 12] have been considered. Then, a procedure is introduced to design a recurrent SNN comprised of leaky integrate and fire (LIF) neurons, effectively establishing a neuromorphic counterpart of the Kalman filter (KF) which is a fundamental and popular strategy for linear deterministic systems [9]. Besides, the KF has a wide application in the areas such as radar signal processing [13, 14], spacecraft state estimation [15, 16], AAM applications such as vision-based navigation [17, 18], and fault-identification in autonomous systems [19]. To further advance the objectives of this study, the proposed framework is to propose a robust estimation framework based on the sliding innovation filter (SIF) algorithm [20], along with a gain modification technique proposed in [21]. The SIF augmented with a gain modification technique serves as a robust filtering strategy that can be considered as a sliding mode estimator of the variable structure filters (VSF) for linear systems and the smooth variable structure filter for nonlinear systems [22, 23]. Moreover, to evaluate the effectiveness of the proposed frameworks for neuromorphic filtering, a comparative analysis was conducted against the original algorithmic frameworks, utilizing a workbench problem with diverse types of uncertainties. Finally, the simulation results demonstrated the acceptable performance of the proposed framework in comparison to their original algorithmic counterparts. It is noteworthy that this research's primary contribution lies in the development of neuromorphic frameworks for optimal and robust filtering of linear dynamical systems. By leveraging the inherent advantages of neuromorphic computing, this study contributes to advancing the field of state estimation in autonomous systems with a focus on reliability, accuracy, and power efficiency.

## II. Preliminaries

This section begins by laying the groundwork with a presentation of some underlying preliminaries. Subsequently, the contributions of this research are outlined. In this study, the linear dynamical systems and the measurement package are assumed to adhere to the following specifications:

$$\dot{x} = Ax + Bu + w \quad (1)$$
$$z = Cx + v \quad (2)$$

where $x$ is the $n_x$-dimensional state vector, $u$ is the $n_u$-dimensional control input vector, $z$ is the $n_z$-dimensional measurement vector, and $A \in R^{n_x \times n_x}$, and $B \in R^{n_x \times n_u}$ refer to the dynamic transition matrix and the input matrix respectively. $C \in R^{n_z \times n_x}$ is the measurement matrix. $w$ and $v$ are the zero-mean Gaussian white noises with the covariance matrices of $Q$, and $R$ respectively.

### A. Continuous time Kalman filtering

The time-continuous form of the Kalman filtering strategy for linear deterministic can be considered as below [24]:

$$\dot{\hat{x}} = A\hat{x} + Bu + K_{KF}(z - \hat{z}) \quad (3)$$

where:
$$\hat{z} = C\hat{x} \quad (4)$$
$$K_{KF} = PC^T R^{-1} \quad (5)$$
$$\dot{P} = AP + PA^T + Q - PC^T R^{-1} CP \quad (6)$$

In the above equations, the symbol ˆ denotes that the parameter has been estimated. $P$ and $K_{KF}$ refer to the state estimation error covariance matrix and Kalman filter gain, respectively.

### B. Robust filtering strategy

To achieve the objective of this study, which is the robust filtering of linear dynamical systems, the sliding innovation filter (SIF) has been considered. The SIF framework closely resembles the previously introduced KF discussed in the previous section, with the primary distinction lying in the formulation of the gain. In order to avoid redundant equations, we present only the gain formulation of the SIF below [20]:

$$K_{SIF} = C^+ sat(|z - \hat{z}|/\delta) \quad (7)$$

where $C^+$ represents the pseudo inverse of the measurement matrix, and $\delta$ refers to the sliding boundary layer which can be tuned by trial and error. However, for the purpose of this research, we adopt the gain formulation proposed in [21] for the SIF filtering strategy. Hence, the following expression is employed:

$$K_{MSIF} = C^+ sat(diag(P^{zz})/\delta) \quad (8)$$

where:
$$P^{zz} = PCP^T + R \quad (9)$$

In the above expression, $P^{zz}$ refers to the innovation covariance matrix.

### C. Neuromorphic computing

In order to develop a neuromorphic estimation framework suitable for implementation on a neuromorphic computer chip, it is necessary to design a network comprising spiking neurons that can emulate the implementation of the estimators' dynamics. Therefore, this section will present a concise overview of the fundamental aspects involved in designing a network of recurrent leaky integrate-and-fire (LIF) neurons that possesses the ability to replicate the dynamic of fully observable linear dynamical systems. The network of LIF neurons can be defined by the following equation [12]:

$$\dot{v} = -\lambda v + Fu(t) + \Omega_s r + \Omega_f s + \eta \quad (10)$$

where, $v \in R^N$ is the vector of neurons membrane potential, $\lambda$ is a decay term, and $F \in R^{N \times n_u}$ matrix, which encodes the input vector. $\Omega_s \in R^{N \times N}$, and $\Omega_f \in R^{N \times N}$ are the slow and fast synaptic connections, $\eta$ is the noise considered on the network, $r \in R^N$ refers to the filtered spike trains which have slower dynamics in comparison to the $s \in R^N$ that is the emitted spike train of the neurons in each time step. The dynamics of the filtered spike trains are provided below.

$$\dot{r} = -\lambda r + s \quad (11)$$

Based on the theory of spike coding network (SCN) [11, 12], a recurrent SNN of LIF neurons can be demonstrated for tracking the $n_x$-dimensional state vector $x$ optimally. This tracking capability relies on two assumptions. First, the estimated state $\hat{x}$ can be decoded from the neural activity using the following linear decoding rule:

$$\hat{x} = Dr \quad (12)$$

where, $D \in R^{n_x \times N}$ is the random fixed decoding matrix that contains the neuron output weights. Second, the neurons have to spike only when their spike reduces the network coding error. To this aim, the firing rule of $\|x - Dr\|_2^2 > \|x - Dr - D_i\|_2^2$ has to be satisfied for each neuron to emit a spike. $D_i$ is $i^{th}$ column of the matrix $D$ which could be considered as $i^{th}$ neuron output kernel that could reflect the change in the error due to a spike of $i^{th}$ neuron. This firing rule ensures that each neuron emits a spike only when it contributes to reducing the predicted error. Thus, by considering these assumptions a recurrent network of LIF neurons capable of implementing a linear dynamical system expressed in Eq. (1) can be defined by the following expression [12]:

$$\dot{v} = -\lambda v + D^T(\dot{x} + \lambda x) - D^T D s \quad (13)$$

The above equation represents the dynamics of the networks' membrane potential that can be used to implement any linear dynamical system.

### D. SNN-based optimal estimation

The focus of this section is on the development of a network of LIF neurons that will be able to implement the linear dynamics of previously introduced KF presented in Eq. (3), using the expression in Eq. (13). Thus, substituting the dynamics of the KF into the Eq. (13) results in the following expression:

$$\dot{v} = -\lambda v + D^T \left( (A\hat{x} + Bu + K_{KF}(z - \hat{z})) + \lambda \hat{x} \right) - D^T D s + \eta \quad (14)$$

Further simplification leads to the final form resembling Equation (10) with additional terms:

$$\dot{v} = -\lambda v + Fu(t) + \Omega_s r + \Omega_f s + \Omega_k r + F_k z + \eta \quad (15)$$

where:

$$F = D^T B \quad (16)$$
$$\Omega_s = D^T(A + \lambda I)D \quad (17)$$
$$\Omega_f = -D^T D \quad (18)$$
$$\Omega_k = -D^T(PC^T R^{-1})CD \quad (19)$$
$$F_k = D^T(PC^T R^{-1}) \quad (20)$$

Here, $F$, $\Omega_s$, $\Omega_f$, $\Omega_k$, and $F_k$ are network weight matrices defined by Eq. (16) to Eq. (20). These matrices are designed to capture the specific dynamics of the KF model and enable the estimation of a fully observable linear dynamical system with partial measurement of the states. It is important to note that updating the matrix $P$ in Eq. (19) and Eq. (20) requires implementing the dynamics expressed in Eq. (6). By examining Eq. (15), it becomes evident that the first term represents the leak term on the membrane potential of neurons, while the subsequent terms contribute to the network connections, reflecting the network's functionality in estimating the linear dynamics of the system. The first term is the potential leak, the next three terms primarily contribute to the *a-priori* prediction phase of the estimation process, and the subsequent two terms, influenced by $\Omega_k$, and $F_k$, are responsible for the connections that work on the *measurement-update* or *a-posteriori* phase of the estimation. The final term represents the introduced noise in the network which imparts stochasticity to the neural activity, akin to the characteristics observed in biological neural circuits. Finally, considering the assumptions introduced in the previous section, in the proposed network, the neurons emit spikes when their membrane potentials reach their respective thresholds, defined as $T_i = (D_i^T D_i)/2$. By adhering to this firing criterion, the network incorporates the spiking behavior of neurons in the estimation process. To extract the estimated states of the considered dynamical system from the neural activity, Eq. (12) can be employed. This equation provides the estimated states from the spiking neural activity within the network. Fig. 1, presents the schematic of the algorithmic KF and its neuromorphic counterpart SNN-KF. Fig. 1(a) illustrates the sequential computation approach employed in the traditional KF as implemented on von Neumann computers. Fig. 1(b) presents the corresponding schematic for SNN-KF, showcasing how the sequential computation has changed into the parallelized computation in the presented network.

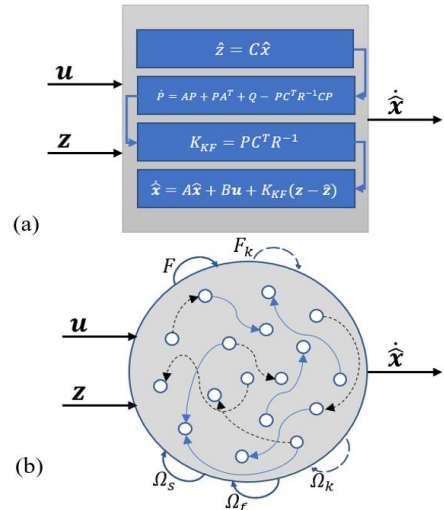

Fig. 1. Schematic diagram for sequential KF and SNN-based KF, (a) sequential KF, (b) SNN-based KF

This section of the paper demonstrates the significant impact of the presented method. By utilizing such approaches, we gain the ability to seamlessly translate our conventional frameworks into SNNs. This translation allows for the implementation of equivalent tasks on neuromorphic chips with enhanced reliability and orders of magnitude lower energy consumption compared to traditional computers. The successful integration of traditional frameworks into SNNs paves the way for leveraging the unique characteristics of neuromorphic chips, enabling substantial energy savings and improved overall performance.

*E. SNN-based robust estimation*

In the preceding section, the SNN-based framework for optimal estimation of linear dynamical systems, built upon on the principles of the KF has been proposed. In this section, we aim to extend this framework to a robust version, capable of handling linear systems affected by bounded uncertainties. To achieve this, we will reformulate Eq. (19), and Eq. (20) using the introduced gain formulation of the SIF as presented in Eq. (8).

$$\Omega_k = -D^T(C^+ sat(diag(P^{zz})/\delta))CD \quad (21)$$
$$F_k = D^T(C^+ sat(diag(P^{zz})/\delta)) \quad (22)$$

Presenting the robust version of the framework is essential as it equips the SNN-based system with the ability to handle and mitigate uncertainties, thereby broadening its applicability to a wider range of real-world scenarios. The gain formulation considered in this study is composed of the measurement matrix, and the diagonal elements of the innovation covariance matrix. This formulation incorporates second-order information regarding the variations in the innovation vector, rendering it faster and more sensitive to the changes in the innovation vector. As a result, this formulation exhibits enhanced robustness compared to its original version introduced in Eq. (7). Furthermore, the utilization of this gain formulation ensures estimation stability in the presence of uncertainties [11]. To update the innovation covariance matrix in each time step of integration, Eq. (9), and Eq. (6) can be implemented. However, to eliminate the need for covariance updates in every time step, the following formulations for $\Omega_k$, and $F_k$ can be employed.

$$\Omega_k = -D^T(C^+ sat(|z - \hat{z}|/\delta))CD \quad (23)$$
$$F_k = D^T(C^+ sat(|z - \hat{z}|/\delta)) \quad (24)$$

The above expressions are derived from the original gain formulation of SIF presented in Eq. (7). By utilizing these equations, we can effectively update the relevant matrices in the network and enhance its performance in the presence of uncertainties. The incorporation of these updated equations represents a valuable contribution to the SNN-based estimation framework. The derived expressions enable a more efficient estimation framework with less computational burden while still maintaining the robustness and stability of the system. However, as discovered in our previous work [21], it is worth noting that there may be a slight degradation in accuracy and increased sensitivity to abrupt changes in the system's states. The utilization of the modified gain formulation and the derived equations contributes to the overall advancement of the SNN-based framework for optimal estimation of linear dynamical systems, especially when computational efficiency is a dominating factor in implementing such frameworks. All the simulations in this paper have been done based on Eq. (21), and Eq. (22).

### III. NUMERICAL SIMULATION

In order to evaluate the performance of the proposed frameworks, they have been applied to a linear system in two distinct conditions. First, a normal condition without uncertainty has been considered, where the frameworks were employed to estimate the states of a dynamical system in comparison to their algorithmic counterparts. This analysis allowed for a direct comparison of the performance between the SNN-based frameworks and traditional algorithmic approaches in a controlled setting to evaluate the applicability of the proposed methods. Subsequently, the SNN-based frameworks are applied to a more realistic problem considering uncertainties in systems' modeling which is inevitable in practice and neuron loss in the network for assessing the effectiveness of the proposed frameworks in practical scenarios. By investigating the results, the performance, reliability, and adaptability of the SNN-based frameworks were thoroughly assessed.

*A. Linear workbench problem*

This section provides the results for the linear system introduced by the following equations:

$$\dot{x} = \begin{bmatrix} 0 & 0 \\ 0 & 1 \end{bmatrix} x + \begin{bmatrix} 0 \\ 1 \end{bmatrix} u + w \quad (25)$$
$$z = \begin{bmatrix} 1 & 0 \end{bmatrix} x + v \quad (26)$$

where:
$$u = -K_c x \quad (27)$$

Simulations have been conducted to evaluate the performance of the proposed framework. The simulations were performed using the parameters provided in TABLE 1, with a total duration of 10 seconds and a time step of 0.01. Initially, the system is simulated without any uncertainties, followed by simulations with the inclusion of uncertainties. It is important to note that during the simulations in this section, the decoding matrix *D* is defined using random samples from a zero-mean Gaussian distribution with a variance of 0.25. This ensured variability in the decoding matrix, reflecting real-world scenarios where the exact decoding matrix might not be known precisely and may exhibit certain uncertainties to be more realistic. The simulations served to assess the robustness and performance of the proposed framework under different conditions, providing valuable insights into its effectiveness in both deterministic and uncertain scenarios. By systematically examining the impact of uncertainties on the estimation process, we gain a comprehensive understanding of the framework's capabilities and limitations in practical applications.

TABLE 1
LINEAR SYSTEM SIMULATION PARAMETERS

| Parameter | Value |
| --- | --- |
| $x_0$ | [10,1] |
| $\hat{x}_0$ | [10,1] |
| $K_c$ | [1, 1.7321] |
| $Q$ | $I/1000$ |
| $R$ | $1/100$ |
| $N$ | 300 |
| $\lambda$ | 0.5 |
| $\delta_{MSIF}$ | 0.005 |

Fig. 2 shows the results obtained from different applications of different filtering strategies. In Fig. 2. (a), the performance of the proposed SNN-based filters in tracking the state $x_1$ is compared. It can be observed that the SNN-based filters have effectively tracked the true state, as demonstrated by the closely aligned tracking curves. The onset figure provides a closer view of the tracking performance, revealing an acceptable error for the SNN-based methods when compared to KF, and MSIF approaches. Similarly, Fig. 2. (b) compares the tracking performance of the filters for the state $x_2$. Once again, the SNN-based strategies have accurately tracked the true state. The onset figure shows the tracking performance of the SNN-based methods has achieved an acceptable range of error in comparison to KF and MSIF. In addition, the onset figure highlights the superior tracking error of SNN-MSIF method compared to the SNN-KF method, further underscoring the effectiveness of the SNN-MSIF approach in accurately estimating the state variables.

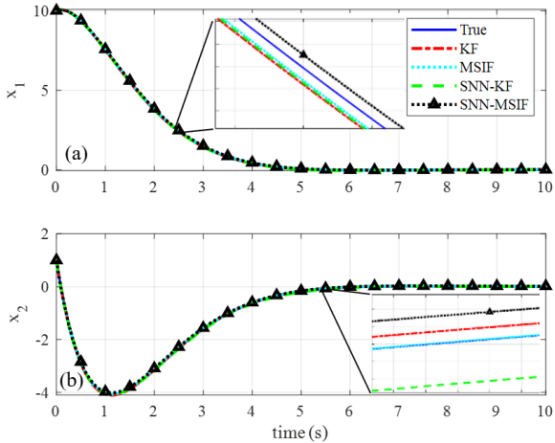

Fig. 2. Time-histories of the true and estimated states obtained from different filters, (a) for the state $x_1$, (b) for the state $x_2$

Fig. 3 depicts the time histories of estimation error $x_i - \hat{x}_i$ within the $\pm 3\sigma$ bounds, which is a criterion for assessing estimation stability. Here, $\sigma_i$ represents the $i$th diagonal element of the covariance matrix $P$. In Fig. 3(a), the estimation error for state $x_1$ is plotted within its respective bound. It can be observed that even in the time duration between 1s to 4s in which the errors are in their worst condition, they have never exceeded the bound. This observation confirms that all the filtering strategies employed in this simulation successfully maintained the estimation stability for the state $x_1$. In addition, the temporal variation of $\pm 3\sigma$ bounds for this state initially set at $\pm 1$ at time $t = 0$, exhibits a monotonically decreasing pattern, indicating a decreasing range of the estimation error and dynamic stability. Similarly, Fig. 3(b) illustrates the estimation stability for state $x_2$. The plot also shows narrower $\pm 3\sigma$ bounds for this state, indicating a more accurate estimation for $x_2$. Fig. 4 presents the comparison of the time histories of root mean squared error (RMSE) obtained from different filtering methods for 100 Monte-Carlo simulations. In Fig. 4(a), the results for state $x_1$ are presented, revealing the superior performance of the SNN-MSIF in comparison to SNN-KF in terms of estimation accuracy throughout the entire time history until $t = 7$s. Between $t = 7$s and $t = 9$s, both SNN-MSIF and SNN-KF demonstrate nearly identical accuracy. However, during the last second of the simulation, SNN-KF exhibits a more accurate estimation. Fig. 4(b) presents the RMSE time histories for the state $x_2$. The comparison indicates that the SNN-KF achieves better accuracy during the time history preceding $t = 3$s. However, for the time history between $t = 3$ and $t = 6$, the SNN-MSIF outperforms SNN-KF by providing a more accurate estimation. Beyond this interval, the estimation error obtained from SNN-MSIF exceeds that of SNN-KF. Overall, it can be inferred that both the SNN-based methods exhibit similar estimation accuracy levels for state $x_2$, especially considering that there are no corresponding measurements available for this state. Importantly, the presented comparison also demonstrates that the SNN-based methods and traditional KF and MSIF approaches operate within the same order of magnitude in terms of estimation accuracy. The chattering which has been observed in the errors for the time duration between $t = 0$s and $t = 1$s can be attributed to the high rate of spiking of the neuron spiking within the networks before

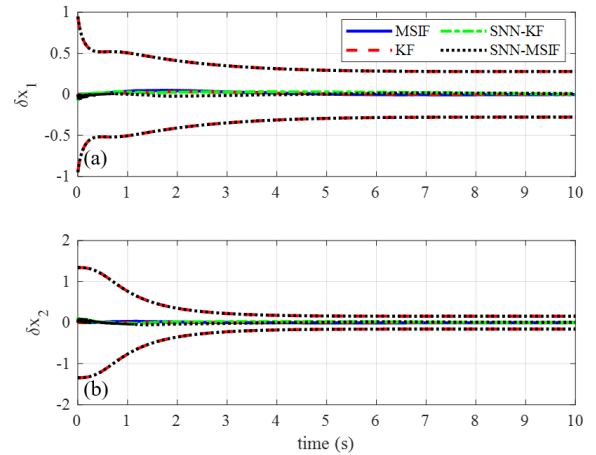

Fig. 3. Time-histories of the estimation errors inside $3\sigma$ bounds obtained from different filters, (a) for the state $x_1$, (b) for the state $x_2$

achieving stability in inter-neuronal communication between their neurons to perform the computations. Further analysis of the spiking patterns will be investigated in the subsequent sections to gain deeper insights into this phenomenon.

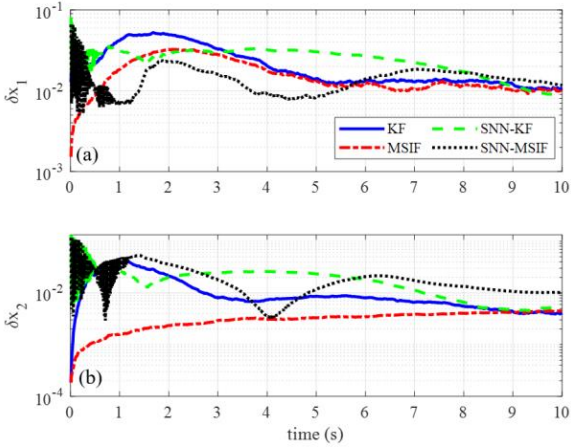

Fig. 4. Comparison of RMSE time-histories obtained from different filters, (a) for the state $x_1$, (b) for the state $x_2$

Moreover, to conduct a quantitative assessment of the estimation accuracies achieved by different filters, the average errors are summarized in the TABLE 2. These results further validate the earlier observations regarding the acceptable accuracy attained by the proposed methods.

TABLE 2
AVERAGE RMSE FOR DIFFERENT FILTERS

| State | KF | MSIF | SNN-KF | SNN-MSIF |
|---|---|---|---|---|
| $x_1$ | 0.0224 | 0.0161 | 0.0248 | 0.0142 |
| $x_2$ | 0.0114 | 0.0031 | 0.0183 | 0.0197 |

Furthermore, to investigate the robustness of the proposed method in the presence of uncertainties, parametric uncertainty has been introduced into the linear system model. To incorporate this type of uncertainty, 10% error is applied to the dynamic matrix $A$ of the model presented in Eq. (1) that is used in designing the proposed recurrent SNN weights, and also in the models used for algorithmic filtering strategies. Then, the Monte-Carlo simulations have repeated under this new condition. Fig. 5 shows the results obtained from different filters for tracking the true state in the presence of uncertainty. In Fig. 5(a), the results for the state $x_1$ are presented, demonstrating effective tracking by all the filters. However, the initial portion of the figure reveals that in the presence of the aforementioned uncertainty, both the MSIF and SNN-MSIF exhibit superior tracking performance with a smaller error compared to KF and SNN-KF, respectively. Fig. 5(b) shows that the KF and SNN-KF deviate from the true state trajectory between $t = 1$s and $t = 5$s. In contrast, both the MSIF and SNN-MSIF track the true states accurately without any deviation. Moreover, the SNN-MSIF demonstrates comparable tracking performance to its counterpart, the MSIF.

These observations validate the robustness of the proposed method in presence of uncertainty.

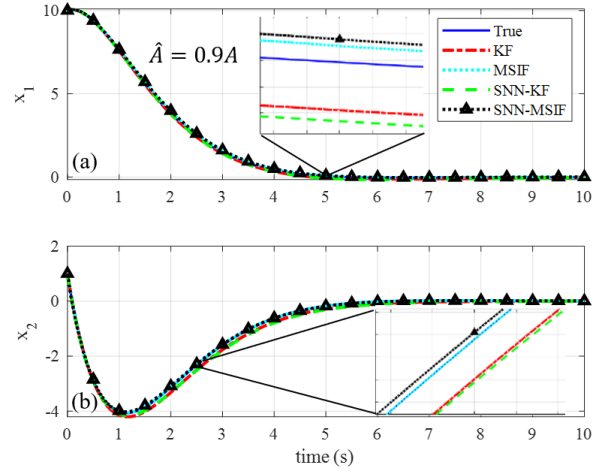

Fig. 5. Time-histories of the true and estimated states obtained from different filters for uncertain case, (a) for the state $x_1$, (b) for the state $x_2$

To streamline the presentation and avoid excessive figures, the estimation error within $\pm 3\sigma$ bound plots is omitted for this and subsequent simulations in this section. Fig. 6 illustrates the temporal variation of RMSEs obtained by filtering strategies in the presence of uncertainty. In Fig. 6(a), the superior accuracy of KF and SNN-KF is observed for the time duration prior to $t = 4$s. Conversely, the MSIF and SNN-MSIF exhibit superior performance in terms of accuracy for the state $x_1$ throughout the remaining simulation time. Fig. 6(b) shows the error variations for the state $x_2$. It is demonstrated that the MSIF and SNN-MSIF consistently achieve more accurate estimation compared to the other methods across the majority of the time history. However, at $t = 9.5$s, they converge to a similar error range. These results provide further confirmation of the robustness of the proposed method in the presence of uncertainty.

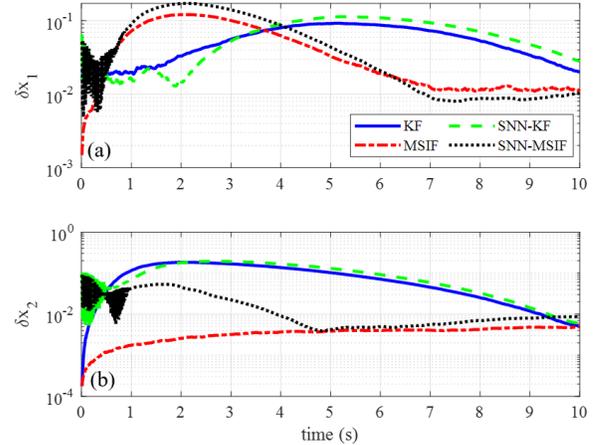

Fig. 6. Comparison of RMSE time-histories obtained from different filters for uncertain case, (a) for the state $x_1$, (b) for the state $x_2$

TABLE 3
AVERAGE RMSE FOR DIFFERENT FILTERS –
UNCERTAIN CASE

| State | KF | MSIF | SNN-KF | SNN-MSIF |
|---|---|---|---|---|
| $x_1$ | 0.0550 | 0.0451 | 0.0639 | 0.0591 |
| $x_2$ | 0.0860 | 0.0035 | 0.0962 | 0.0171 |

Furthermore, the average errors for this case have been tabulated in TABLE 3. The results reaffirm the superior performance of the MSIF and SNN-MSIF compared to the KF-based methods in the presence of uncertainties. Thus, these findings provide strong evidence for the robustness of the proposed framework in handling uncertainties, which are inherent and inevitable in real-world problems.

*B. Sensitivity to number of neurons*

To evaluate the ability of the designed network under damaged conditions, where a loss of neurons occurs, Monte-Carlo simulations have been repeated multiple times with three different neuron counts ranging from $N = 450$ to $N = 50$, reducing the number of neurons by 50 each time. These experiments aimed to assess the sensitivity of the proposed estimators to number neurons. The results of the simulations revealed the impact of neuron count on the performance of the estimators. When the number of neurons exceeded a certain threshold, the network estimators remained stable, as indicated by the means of $\pm 3\sigma$ criterion. However, a significant increase in high-frequency chattering has been observed in the estimation of the true states. On the other hand, when the number of neurons was reduced below a certain number, the proposed estimators lost their ability to accurately track the true state and exhibited instability. Fig. 7 presents the estimation result for the case of $N = 450$. In Fig. 7(a), the estimation of state $x_1$ is presented, indicating that the framework achieved a near-tracking of the true state. However, the onset figure reveals the presence of a high frequency chattering in the estimated state for the SNN-based frameworks.

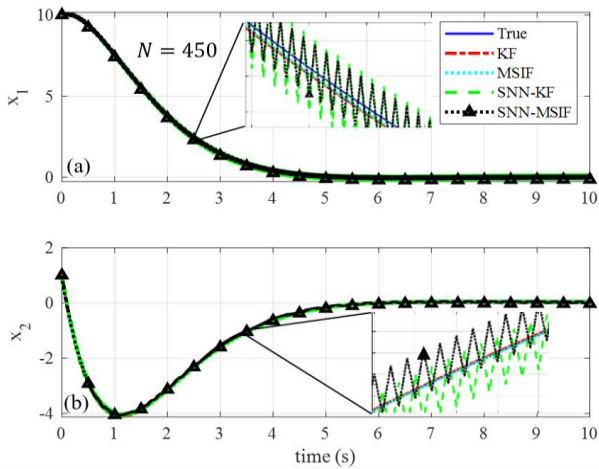

Fig. 7. Time-histories of the true and estimated states obtained from different filters for $N = 450$, (a) for the state $x_1$, (b) for the state $x_2$.

Similarly, Fig. 7(b) presents the estimation of the state $x_2$ obtained from different filters. The results demonstrate that the proposed frameworks almost tracked the true state, but the onset figure also reveals the presence of high frequency chattering in the estimated states. These findings highlight the sensitivity of the proposed estimators to the number of neurons in the network and emphasize the need to carefully consider the neuron count to ensure stable and accurate estimations.

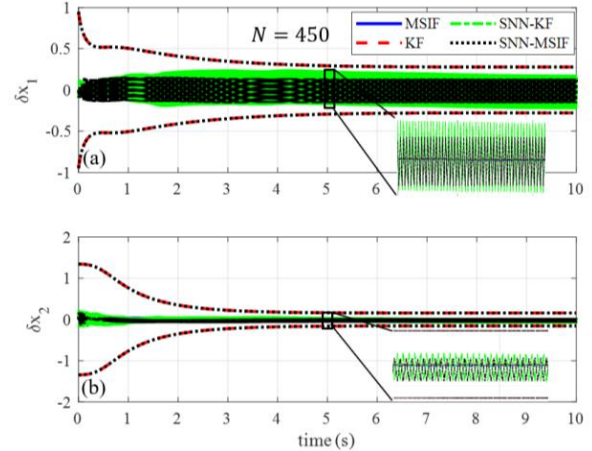

Fig. 8. Time-histories of the estimation errors inside $3\sigma$ bounds obtained from different filters for $N = 450$, (a) for the state $x_1$, (b) for the state $x_2$

Fig. 8 presents the estimation error within $\pm 3\sigma$ bounds for the case of $N = 450$. In Fig. 8(a), the results depict the estimation error for state $x_1$. Notably, the estimation error has not exceeded the $\pm 3\sigma$ bounds. Thus, considering this criterion, the proposed SNN-based frameworks have performed a stable estimation. However, the inset figure reveals the presence of chattering in error, which corresponds to the chattering observed in the state estimation. Similarly, Fig. 8(b) presents the estimation error for the state $x_2$, demonstrating a consistent pattern However, the obtained results for the state $x_2$ has a smaller domain for the chattering of the errors, with a narrower $\pm 3\sigma$ bounds compared to state $x_1$. To avoid the excessive number of figures, we have omitted the simulation results for all the different numbers of neurons in this section. However, it is worth noting that when considering the number of neurons between $N = 100$ and $N = 350$, the estimation accuracies were still acceptable in comparison to our normal case with $N = 300$ despite the degradation caused by the loss of neurons. In contrast, for the case of $N = 50$ the SNN-KF framework nearly lost its ability to estimate true states and became completely unstable. However, for the SNN-MSIF, although its estimate is deviated from the true state, it remained relatively stable. Fig. 9 presents the results obtained for tracking the true states using different filters for the case $N = 50$. In Fig. 9(a), it is evident that the SNN-KF has deviated from the true state for the state $x_1$. On the other hand, the SNN-MSIF has tracked the true state with a smaller deviation from the true states compared to SNN-KF, indicating a more robust estimation capability in the face of neuron loss. Fig. 9(b), shows the estimation results obtained for the tracking of the state $x_2$. It demonstrates that the SKK-KF has completely deviated in tracking the true state from $t = 0$s until $t = 7$s, but thereafter, it managed to closely track the true state. However, in comparison to SNN-KF, the figure indicates an acceptable tracking performance for almost the entire simulation duration. Therefore, the SNN-MSIF exhibits superior resilience to neuron loss when compared to the SNN-KF, as demonstrated by the tracking performance in the presence of $N = 50$.

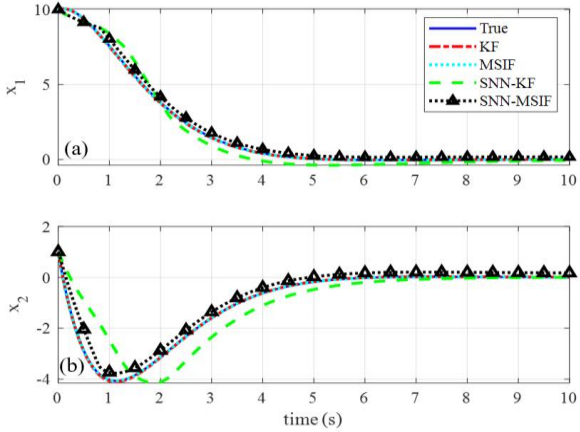

Fig. 9. Time-histories of the true and estimated states obtained from different filters for $N = 50$, (a) for the state $x_1$, (b) for the state $x_2$

Fig. 10 shows the $\pm 3\sigma$ plots for the simulation case with $N = 50$. In Fig. 10(a), the results obtained for the state $x_1$ are presented the. It is demonstrated that the stability of the proposed SNN-based methods has been compromised due to the loss of neurons. Here, the SNN-KF becomes completely unstable until $t = 7$s, after which the estimation error has converged to zero. On the other hand, the SNN-MSIF exhibits an initial large error prior to $t = 1$s, but gradually decreases thereafter, and remains constant after $t = 5$s for the rest of the simulation time. In Fig. 10(b), the simulation result for the state $x_2$ are depicted It is demonstrated that the estimation error obtained from SNN-KF experiences a drastic increase at the beginning and becomes completely unstable until $t = 6$s after which the error returns within the specified bounds. In contrast, the SNN-MSIF exhibits a relatively stable estimation where the error never exceeds the bounds, although it remains on the boundary throughout the entire simulation duration. From these results, it can be concluded that the SNN-MSIF exhibits greater robustness in the presence of neuron loss compared to the SNN-KF. The SNN-MSIF demonstrates stable estimation, and the error remains on the boundary, while the SNN-KF suffers from instability and large errors during certain time intervals.

To compare the performance of the proposed methods in terms of robustness and accuracy under neuron loss condition, Fig. 11 presents the comparison of the averaged RMSEs obtained for the states $x_1$, and $x_2$ from both the SNN-KF, and SNN-MSIF. The results reveal three distinct regions when varying the number of neurons in the designed network. In region 1, where the number of neurons is less than 100, as previously demonstrated for $N = 50$, the estimation accuracy significantly degrades for both frameworks. This degradation between 100 to 300, the estimation accuracy remains within a satisfactory range compared to the $N = 300$ case in our first can be attributed to the instability of the estimators in this region. In region 2, where the number of neurons varies simulation from the previous section. Also, the obtained results demonstrate a gradual reduction in RMSE by increasing the number of neurons from 100 to 300. For the specific dynamic considered in these simulations, the best performance with the minimum error is achieved when $N = 300$. In region 3, where the number of neurons surpasses 300, as it has been presented previously for the case of $N = 450$, the errors exhibit a significant increase, indicating the occurrence of chattering in the estimated states. Consequently, selecting a number of neurons beyond this region leads to either instability or chattering in the estimation process. Thus, it is demonstrated that for any linear dynamical system, there exists an appropriate region for selecting the number of neurons to achieve acceptable estimation results. Deviating from this region results in either instability or the chattering in the estimation process. The results presented in the Fig. 11 demonstrate that even in region 2, where the neurons are less than 300, the estimation accuracy is degraded when neurons are lost. However, the loss of accuracy remains within an acceptable range for with $N < 300$, and the estimation does not become unstable. This finding confirms the robustness of the proposed network to neuron loss, which is attributed to the inherit scalability of SNNs that makes them more reliable under such conditions. To further investigate the neural activity and gain insight into the network's behavior when instability or chattering occurs in the estimated states, the spiking patterns were examined for a single simulation run with $N = 300$, $N = 50$, and $N = 450$. Fig 12 presents the spiking pattern obtained for the SNN-MSIF in the case of $N = 300$. The result reveals that during the initial 100 time steps of the simulation, before the proposed filter converges to the true state, a significant number of the neurons emit spikes, indicating a regime with a huge neural activity. The excessive neural activity could be a source of the observed chattering in the RMSEs obtained before $t = 1$s in Fig. 4 and Fig. 6. Moreover, the spiking pattern shows that the network performs with only 3.2% of possible neural activity for almost entire simulation duration. After convergence, when the estimation error becomes negligible, the network operates with just 1.5% of possible neural activity. This is a consequence of the event-driven nature of SNNs, in contrast to time-driven nature of traditional ANNs, where all neurons are active in entire simulation time steps, resulting in 100% neural activity. Therefore, it could be concluded that using SNNs instead of traditional ANNs leads to a significant reduction in energy consumption for computations performed in the neural network. As a result, the proposed neural network can be orders of magnitude more energy efficient in comparison to traditional methods. Fig. 13, presents the spiking pattern obtained from SNN-MSIF for the case of $N = 450$. The obtained result demonstrated a huge amount of neural activity which can be considered as a source for chattering in the estimated states.

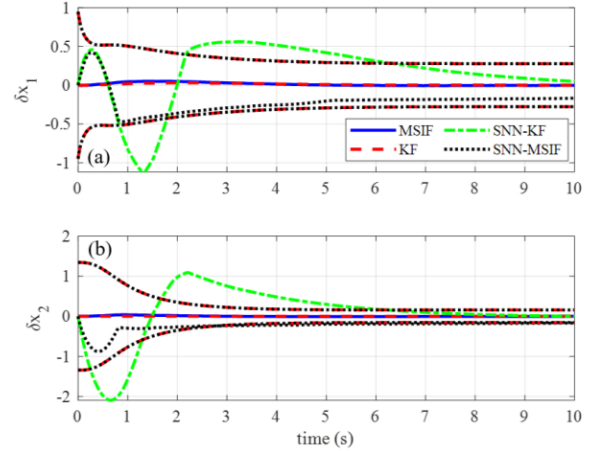

Fig. 10. Time-histories of the estimation errors inside $3\sigma$ bounds obtained from different filters for case $N = 50$, (a) for the state $x_1$, (b) for the state $x_2$

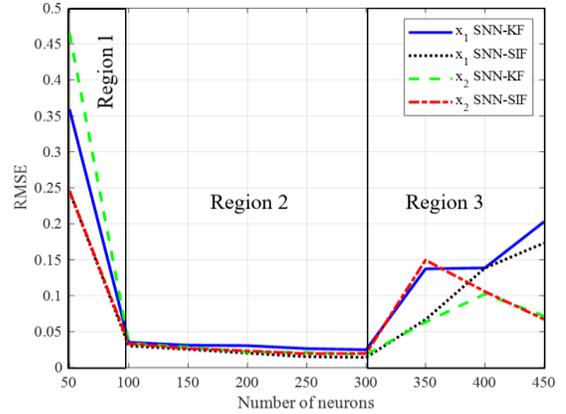

Fig. 11. Averaged RMSEs for the states $x_1$, and $x_2$ obtained from the filtering strategies SNN-KF and SNN-MSIF for various number of neurons.

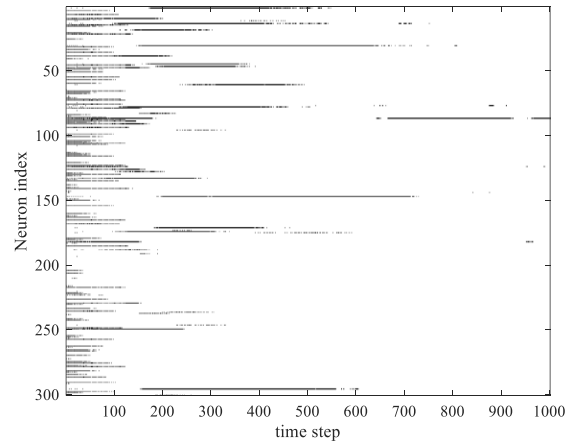

Fig. 12. Spiking pattern obtained for SNN-MSIF for $N = 300$

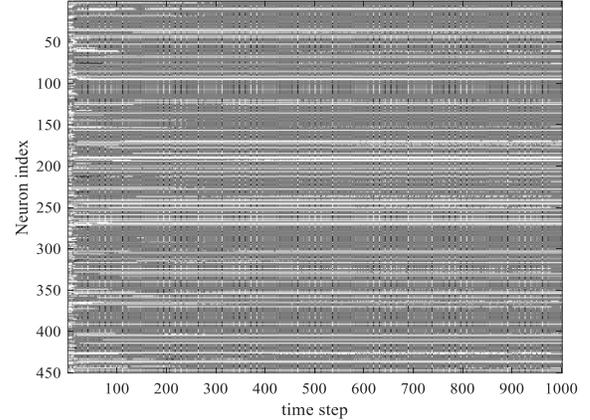

Fig. 13. Spiking pattern obtained for SNN-MSIF in the case of $N = 450$

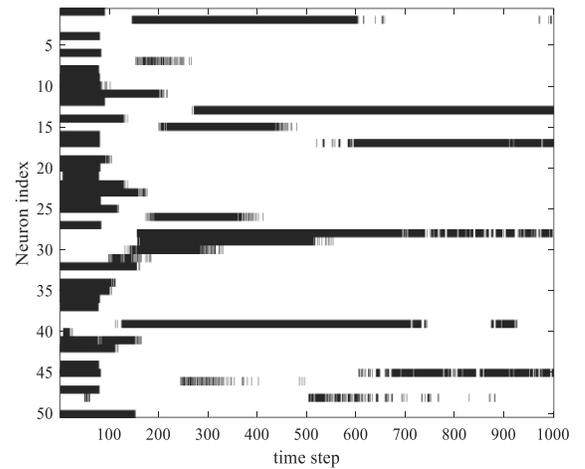

Fig. 14. Spiking pattern obtained for SNN-MSIF in the case of $N = 50$

Fig. 14 presents the spiking pattern for the SNN-MSIF in the case of $N = 50$. The obtained results reveal a minimal amount of neural activity within the network, which is inadequate for the network to be able to perform the precise computations required by the considered system. Thus, as shown in Fig. 9, this limited neural activity leads to a degradation in the estimation accuracy and results that deviate from the true state.

## IV. CONCLUSION

The energy efficiency and reliability of computers in autonomous systems have emerged as two crucial factors in industries such as space robotics, autonomous systems, advanced air mobility, internet-of-things, and swap engineering. In this context, based on spike coding theory, the SNN-KF is proposed for optimal estimation of linear systems. Next, a robust SNN-based method based on MSIF is introduced. Further, to demonstrate the applicability of these methods, they are compared with their counterparts namely,

KF and MSIF. Furthermore, the robustness of the SNN-MSIF compared to SNN-KF has been demonstrated in the presence of uncertainties and neuron loss through Monte-Carlo simulations. Our results consistently demonstrated that the SNN-MSIF outperforms the SNN-KF in terms of accuracy and robustness. To sum up, one of the key advantages of our proposed frameworks lies in their compatibility with neuromorphic computer chips. By leveraging the energy efficiency and scalability of neuromorphic computers, these frameworks offer significant advantages over traditional von Neumann computers when performing equivalent tasks. Implementing the proposed method for concurrent estimation and control of satellite rendezvous, which is a vital maneuver in space robotics for on-orbit servicing, has been considered for the future work of this research.

## V. Conflict of interest statement

The authors declare that they have no conflicts of interest related to this research. The study was conducted in an objective and unbiased manner, and the results presented herein are based on rigorous analysis and interpretation of the data. The authors have no financial or personal relationships with individuals or organizations that could potentially bias the findings or influence the conclusions of this study.